\documentclass[10pt,twocolumn,letterpaper]{article}

\usepackage{cvpr}
\usepackage{times}
\usepackage{epsfig}
\usepackage{graphicx}
\usepackage{amsmath}
\usepackage{amssymb}
\usepackage{float}

\usepackage{algorithm2e}
\makeatletter
\renewcommand{\@algocf@capt@plain}{above}
\makeatother

\usepackage[pagebackref=true,breaklinks=true,letterpaper=true,colorlinks,bookmarks=false]{hyperref}

\cvprfinalcopy 

\def\cvprPaperID{6790} 

\ifcvprfinal\fi
\begin{document}


\title{Adversarial point set registration}

\author{Sergei Divakov\\
Skolkovo Institute of Science and Technology\\
{\tt\small sergei.divakov@skoltech.ru}
\and
Ivan Oseledets\\
Skolkovo Institute of Science and Technology\\
Institute of Numerical Mathematics RAS\\
{\tt\small i.oseledets@skoltech.ru.org}
}

\maketitle

\begin{abstract}
   We present a novel approach to point set registration which is based on one-shot adversarial learning. The idea of the algorithm is inspired by recent successes of generative adversarial networks. Treating the point clouds as three-dimensional probability distributions, we develop a one-shot adversarial optimization procedure, in which we train a critic neural network to distinguish between source and target point sets, while simultaneously learning the parameters of the transformation to trick the critic into confusing the points. In contrast to most existing algorithms for point set registration, ours does not rely on any correspondences between the point clouds. We demonstrate the performance of the algorithm on several challenging benchmarks and compare it to the existing baselines.
\end{abstract}

\section{Introduction}
Point set registration is one of the fundamental problems in 3D computer vision. It is formulated as finding a rigid transformation which would best align two point clouds corresponding to the same object or scene. It is key to such applications as localization of autonomous vehicles \cite{machines5010006} and building accurate maps for complex environments. In industrial automation, the registration problem often arises when a robotic system needs to estimate the pose of an object in order to plan a motion to grasp it - an illustrative example of this is bin picking \cite{Rahardja}.

It is common to distinguish between two different setups of the registration problem: global and local. Local methods produce a tight alignment given that the initial approximation of the transformation is good enough. However, they usually do not work if the point clouds are strongly misaligned. Global methods, on the contrary, do not depend on the initial transformation. They are typically used to produce a rough approximation of the transformation, which can then be refined by a local algorithm.

Most of the currently existing algorithms for point set registration, both global and local, are based on the idea of building correspondences between geometric features or individual points of the point clouds and finding the transform which respects this correspondence. This approach has resulted in a great variety of successful methods \cite{Rusu:2009:FPF:1703435.1703733} \cite{Rusinkiewicz:2001:EVO}. However, in cases when the correspondence is modelled incorrectly such algorithms may fail to work.

Our idea is to apply a method of adversarial learning to solve the registration problem. We build upon the recent successes of generative adversarial networks \cite{NIPS2014_5423} - GANs, which have proven to be very effective in modelling complex multidimensional distributions \cite{DBLP:journals/corr/RadfordMC15} \cite{NIPS2015_5773}. Particularly, in \cite{conneau2017word}, the adversarial learning setup is used to find an orthogonal mapping, which aligns the distributions of word embeddings of two different languages, thus resulting in a translation system built without using parallel corpora.

This inspired us to develop a one-shot adversarial optimization procedure, where the critic neural network is trained to distinguish between points from source and target point sets, while the generator, which in our case is just an $SE(3)$ mapping, trains to fool the critic, thus finding the transform which best aligns them.

Our approach eradicates the need to explicitly match individual points or features. Using a neural network as an approximation for inter-distribution distance between the point sets results in a quasi-linear algorithm which is capable of producing a tight alignment regardless of the quality of initial approximation. The resulting algorithm demonstrates impressive results, outperforming current methods in terms of solution quality and scaling properties. The resulting algorithm demonstrates impressive results, outperforming current methods in terms of solution quality and scaling properties.

The main contributions of the work are as follows:
\begin{itemize}
    \item We describe a novel approach for point set registration, which is based on adversarial learning.
    \item We provide an experimental study of the approach, showing cases of success and failure as well as elaborate on the technical tricks used to make it work.
    \item We measure the effectiveness of our algorithm on a number of illustrative problems and compare it to the existing baselines.
\end{itemize}

\section{Algorithm}

\subsection{General framework}
Given two sets of points $X, \tilde{X} \subset \mathbb{R}^3 $, we need to find a rigid transform $M \in SE(3)$, such that $M[\tilde{X}]$ and $X$ are best aligned. 



Let us introduce two random variables $\xi, \hat{\xi}$.
\begin{align}
    \xi \sim P_X, \\
    \hat{\xi} \sim P_{\tilde{X}},
\end{align}
where $P_X, P_{\tilde{X}}$ are atomic probability densities associated with source and target point sets:
\begin{align}
    P_X(x) = \frac{1}{|X|} \sum_{i = 1}^{|X|} \delta(x - x_i), \\ 
    P_{\tilde{X}}(x) = \frac{1}{|\tilde{X}|} \sum_{i = 1}^{|\tilde{X}|} \delta(x - \tilde{x}_i).
\end{align}
Then, we can reformulate the point set registration problem as a problem of minimization of a divergence $\mathbf{D}$ between these distributions.
\begin{align}
    \min_{M} \ \mathbf{D}(\xi, M[\hat{\xi}]).
\end{align}

This problem formulation is standard for Generative adversarial networks, which have recently proven to excel in modelling complex multidimensional distributions. In GAN setting, an intractable divergence is approximated using a neural network, called the critic, and the optimization is performed via an adversarial game, in which another network - a generator - tries to fool the critic by learning to produce believable samples from the target distribution.

\begin{figure}[H]
    \centering
    \includegraphics[width=\linewidth]{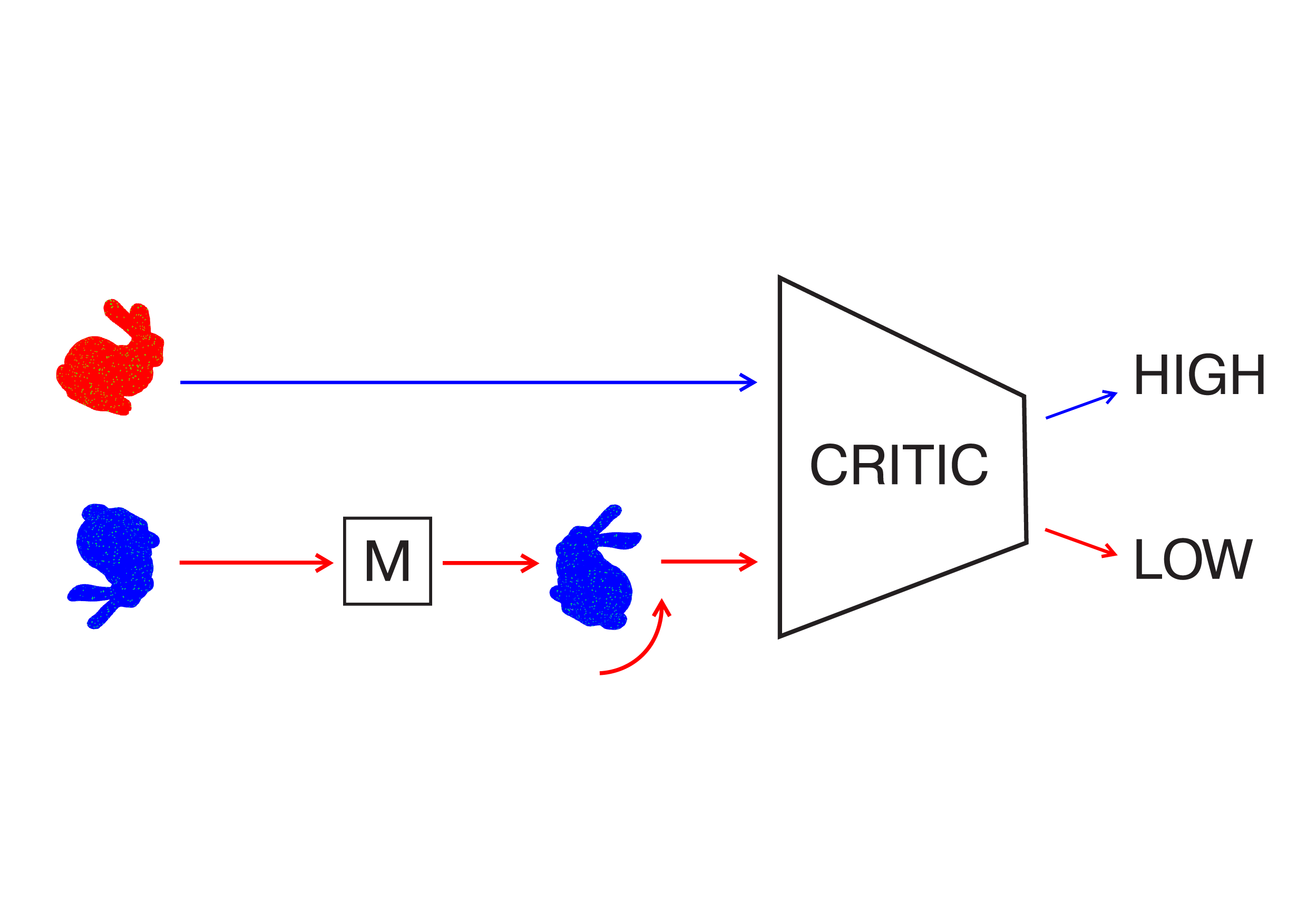}
    \caption{Adversarial setup for point set registration. Little green dots correspond to samples from the target and the source point clouds}
    \label{fig:1}
\end{figure}

In our case, however, the generator isn't trained to sample from the distribution. Rather, it learns a rigid transformation, which, applied to the samples from the source point set, will make them indistinguishable from the target.
In fact, point set registration problem can be viewed as a special case of domain adaptation, which is another example of the successful application of GANs. \cite{Ganin:2016:DTN:2946645.2946704}

\subsection{Metric}
In our work we optimize the Wasserstein-1 metric, also known as Earth-movers distance:
\begin{equation}
\mathbf{D}_{W} [\xi, \hat{\xi}] =
\inf_{\gamma \in \pi (\it{P}, \it{\tilde{P}})} \mathbb{E}_{x, y \sim \gamma} \ d(x, y),
\end{equation}

where $d(x, y)$ is the distance function and the infimum is taken over all joint distributions with marginals equal to $\it{P}$ and $\it\tilde{P}$.

Similarly to \cite{1701.07875}, we solve the optimization, by using the following dual formulation of the W-1 metric
\begin{equation}
    \mathbf{D}_{W} [\xi, \hat{\xi}] = \sup_{||f||_{L} \leq 1}  \mathbb{E}_{x \sim \it{P}} f(x) - \mathbb{E}_{x \sim \it{\tilde{P}}} f(x),
\end{equation}
and approximating the 1-Lipschitz function $f$ with a shallow neural network.

We note that other statistical distances may be used to solve the problem. The Wasserstein metric was chosen because we found that it results in a more robust convergence.

\subsection{Adversarial setting}

Let $f_{\theta_C} : \mathbb{R}^3 \to  \mathbb{R} $ be the mapping performed by the critic neural network. Then the objective of the critic reads:
\begin{align}
    \mathit{L}_C (\theta_C) = - \large{[} \frac{1}{n}\sum_{i = 1}^n f_\theta(x_i) - 
    \frac{1}{m} \sum_{j = 1}^m f_\theta(M(\tilde{x}_j)) \large{]}.
\end{align}
The summations are performed over randomly sampled mini-batches from source and target point clouds.

In order to ensure the Lipschitz property of $f_\theta$ we use gradient penalty \cite{1701.07875}:
\begin{align}
    L_{GP}(\theta_D) = \frac{1}{n} \sum_{i = 1}^n (| \nabla_{\theta_D} f_{\theta_D} (\hat{x}_i) |^2_2 - 1)^2,
\end{align}
where $\hat{x}_i = \alpha x + (1 - \alpha) M(\tilde{x})$ are interpolates obtained by mixing points from target and source point clouds with a randomly sampled coefficient $\alpha \sim \textrm{Uniform} \ [0, 1]$.

Overall, the loss of the critic reads:
\begin{align}
L_C^{full}(\theta_C) = L_C(\theta_C) + \lambda L_{GP}(\theta_C).
\end{align}
Let $\theta_G$ be the parameters of the mapping $M$, then the loss of the generator in this case is just:
\begin{align}
    L_G(\theta_G) = -\frac{1}{m} \sum_{j = 1}^m f_{\theta_C} (M_{\theta_G}(\tilde{x}_j)).
\end{align}

To solve this optimization problem, we use an alternating procedure, sequentially updating the parameters of rotation ("generator") and critic using gradient steps. At each step, we select a batch from the source point cloud and a batch from the target point cloud. Then we apply the currently learned transformation to the source batch, compute the loss function and update the parameters of the model using a stochastic optimization algorithm. \\
The full alignment algorithm is outlined as follows:
\begin{algorithm}[h]
\SetAlgoLined
\KwIn{Source and target point clouds $\tilde{X}, X$}
\KwOut{$\theta_{G}$ - rotation vector and translation }
 $N_{Epochs}$ - the number of epochs to train \\
 $K_{critic}$ - the number of steps to train the critic \\
 $K_{generator}$ - the number of steps to train the generator \\
 \While{epoch $<$ $N_{Epochs}$}{
  \For{$K_{critic}$ steps}
    {
        Sample mini-batch $x$ from $X$\\
        Sample mini-batch $\tilde{x}$ from $\tilde{X}$\\
        Apply transformation $M$ to $\tilde{x}$\\
        Compute $L_{C}^{full}(x, M(\tilde{x}))$\\
        Update parameters of the critic: \\ $\theta_{C} = \theta_{C} - \nabla_{\theta_C} L_{C}^{full}(x, M(\tilde{x}))$\\
    }
  \For{$K_{generator}$ steps}{
        Sample mini-batch $\tilde{x}$ from $\tilde{X}$\\
        Apply transformation $M$ to $\tilde{x}$\\
        Compute $L_{G}(M(\tilde{x}))$\\
        Update parameters of the generator: $\theta_{G} = \theta_{G} - \nabla_{\theta_C} L_{G}(M(\tilde{x}))$\\
   }
 }
 \caption{Adversarial learning algorithm to find the rigid transformation between point clouds}
\end{algorithm}

\subsection{Parametrization}

To parametrize the rotational part of the transformation $M$ we use exponential mapping of $\mathfrak{so}(3)$ Lie algebra, also known as the axis-angle representation. We find it superior to the quaternion representation because it results in an unconstrained optimization procedure well-suited for stochastic gradient descent algorithms.

For the architecture of the critic, we use a shallow neural network with 4 dense layers of size 32 and ReLU activations between them.

\subsection{Learning}

Training generative adversarial networks is often notoriously hard. However, in our case, the optimization procedure seems to be rather stable and usually converges to a plausible result.
We have found that several tricks have a positive influence on the robustness of the algorithm. For cases with significant surface overlap and strong initial misalignment, it helps to normalize the point sets prior to running the optimization.
Standard techniques, such as decaying the learning rate and increasing the batch size help to make the convergence more stable.

We have found that different tasks are best solved with different training parameters, i.e. problems of local matching are solved better by setting a small learning rate for the generator and large batch size, while a rough alignment can be obtained faster by optimizing the generator with a high learning rate.

We also find that when the overlap of the source and the target point sets is small, it helps to search for rotation and translation independently. At first glance, however, it is not obvious how it can be achieved. If we try to first learn the translation between strongly rotated point sets, it is likely that the model will just converge to a difference of means. If, on the other hand, we try to learn the rotation first, without having the correct translation, the model might try to simply maximize the overlap of points, which is not what we need.

\begin{figure}[H]
    \centering
    \includegraphics[width=\linewidth]{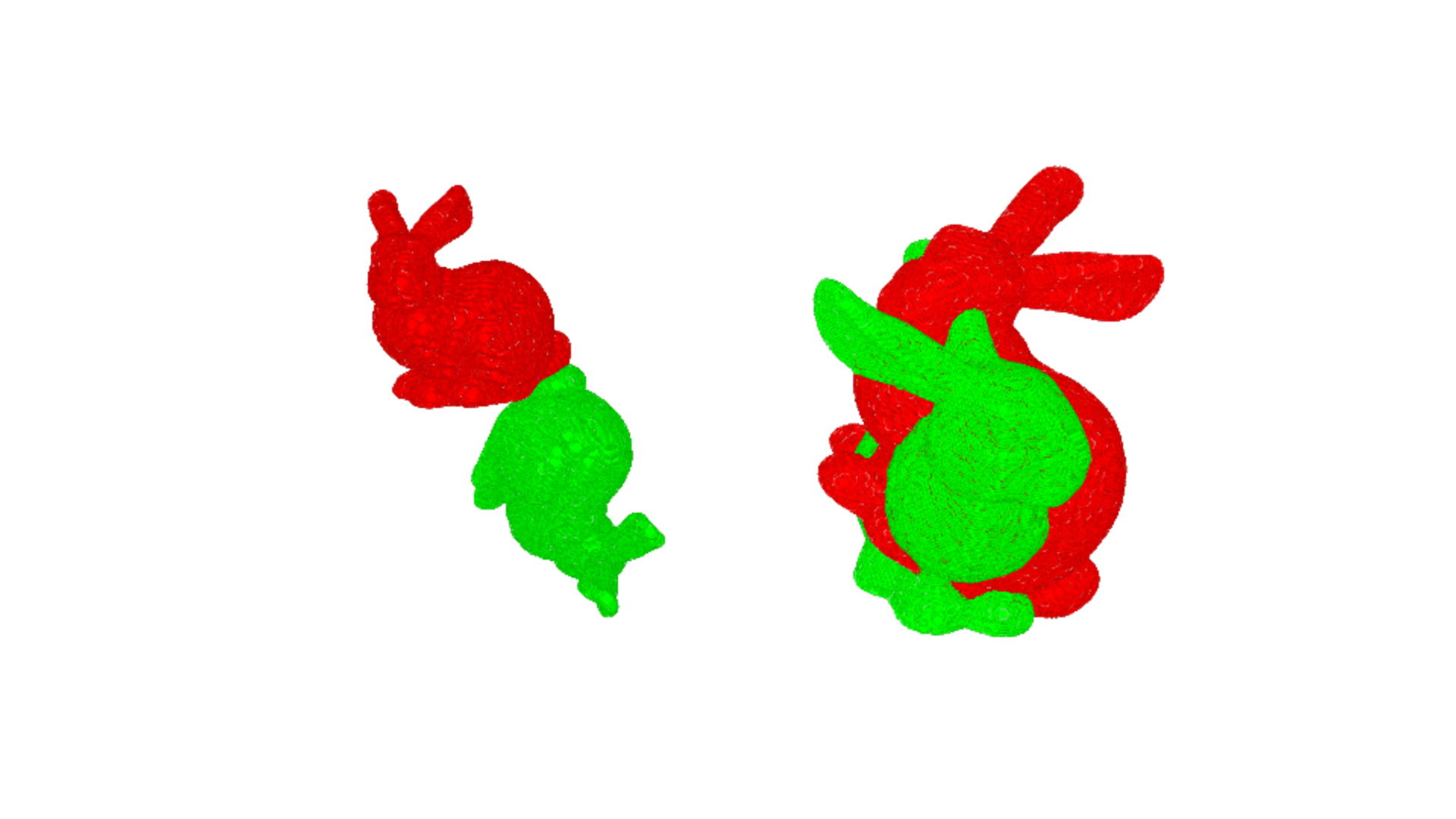}
    \caption{An example when the model converged to a rotation which maximizes the overlap, but is far from the true }
    \label{fig:2}
\end{figure}

Therefore, if we try to learn the rotation first, we need to to make sure that the critic distinguishes points only by angular properties, but not based on the mean or scale of the mini-batch. To achieve this, we use the following trick: on each epoch of training, we shift source and target point sets by different random vectors, sampled from a normal distribution. In addition, we re-scale them by a common random factor. Thus we ensure that the critic learns to distinguish point sets with no regard to scale or absolute positions of points in the mini-batch.

\subsection{Computational complexity}

The complexity of each epoch of training is $O(|X| + |\tilde{X}|)$. The most computationally demanding steps of the algorithm are forward and backward passes of the neural network. However, since the architecture of the discriminator is very simple, the training usually takes a reasonable amount of time. The number of training epochs needed to achieve alignment varies, but is typically in the order of hundreds.

\section{Experiments}

Since the described approach is suitable for both local and global registration settings, we benchmark it both against local and global algorithms.

We have chosen ICP \cite{121791} as a baseline for local registration. In the global registration setting, we compare our method to Fast Global Registration \cite{Zhou2016}. For the implementation of Fast Global Registration, ICP and visualization routines we use Open3D library \cite{Zhou2018}.

\subsection*{\bf{Datasets}}

To compare against local methods, we use Stanford Bunny mesh \cite{Curless:1996:VMB:237170.237269}. We perform a series of augmentations to the model and measure the performance of the algorithms with respect to severity of augmentation.

To demonstrate the performance of our algorithm in the global registration setting, we use our own data, which we make publicly available. It consists of a sparse triangulated mesh of a front part of a car and a dense depth image, acquired from a stereo sensor.

\subsection*{\bf{Evaluation metric}}
Similarly to \cite{Vongkulbhisal_2018_CVPR} as an evaluation metric we chose the angular distance between the obtained rotation and the ground truth. It is computed using Frobenius distance between the rotation matrices \cite{Hartley2013}:
\begin{align}
    d_{ang}(R_{gt}, R) = 2 \arcsin \large{[} ||R_{gt} - R||_{F} / \sqrt{8} \large{]}
\end{align}
We consider registration successful, if the angular distance is less than 4 degrees.

\subsection{\bf{Synthetic data}}

First, we measure the performance of our algorithm in a number of synthetic experiments, using reconstructed meshes of Stanford bunny. In the first series of experiments, we show how our algorithm performs when the translation is fixed to zero. In this setting, only the rotation is optimized.

\subsubsection{\bf{Initial alignment}}

\begin{figure}[H]
    \centering
    \includegraphics[width=\linewidth]{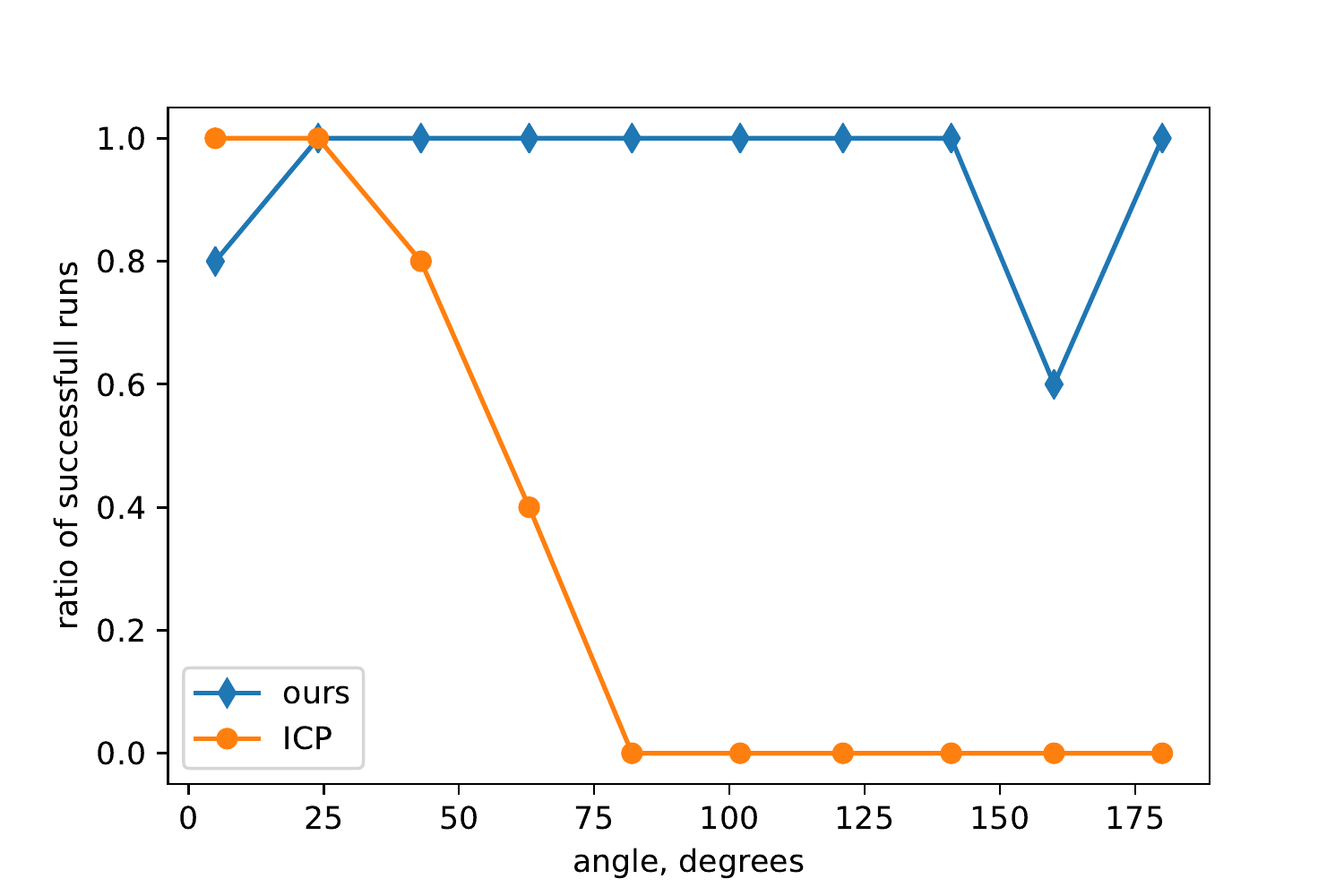}
    \caption{Success ratio to rotation angle}
    \label{fig:3}
\end{figure}

For a number of different magnitudes - from 0 to 180 degrees -  we randomly sample several rotations and measure the success rates of each algorithm.

\subsubsection{\bf{Noise}}

\begin{figure}[H]
    \centering
    \includegraphics[width=\linewidth]{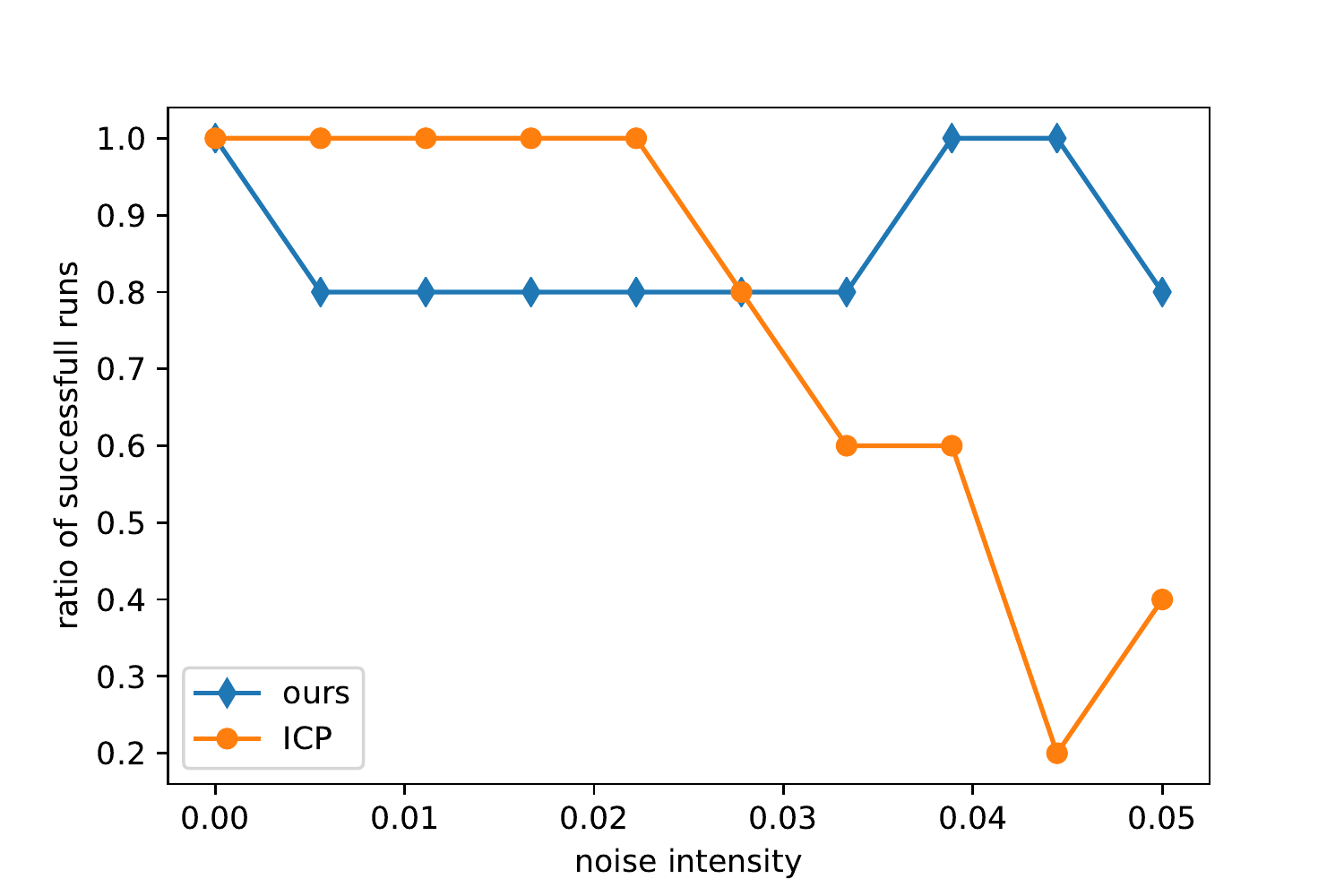}
    \caption{Success ratio to noise intensity}
    \label{fig:4}
\end{figure}

We add Gaussian noise with zero mean and fixed standard deviation to every point of the target and source point sets and measure the quality of the resulting solutions as a function of noise intensity. The magnitude of the initial rotation is fixed to 24 degrees. The value of noise varies between 0.01 and 0.05 multiplied by a standard deviation of the point sets.

\subsubsection{\bf{Partial overlap}}

\begin{figure}[H]
    \centering
    \includegraphics[width=\linewidth]{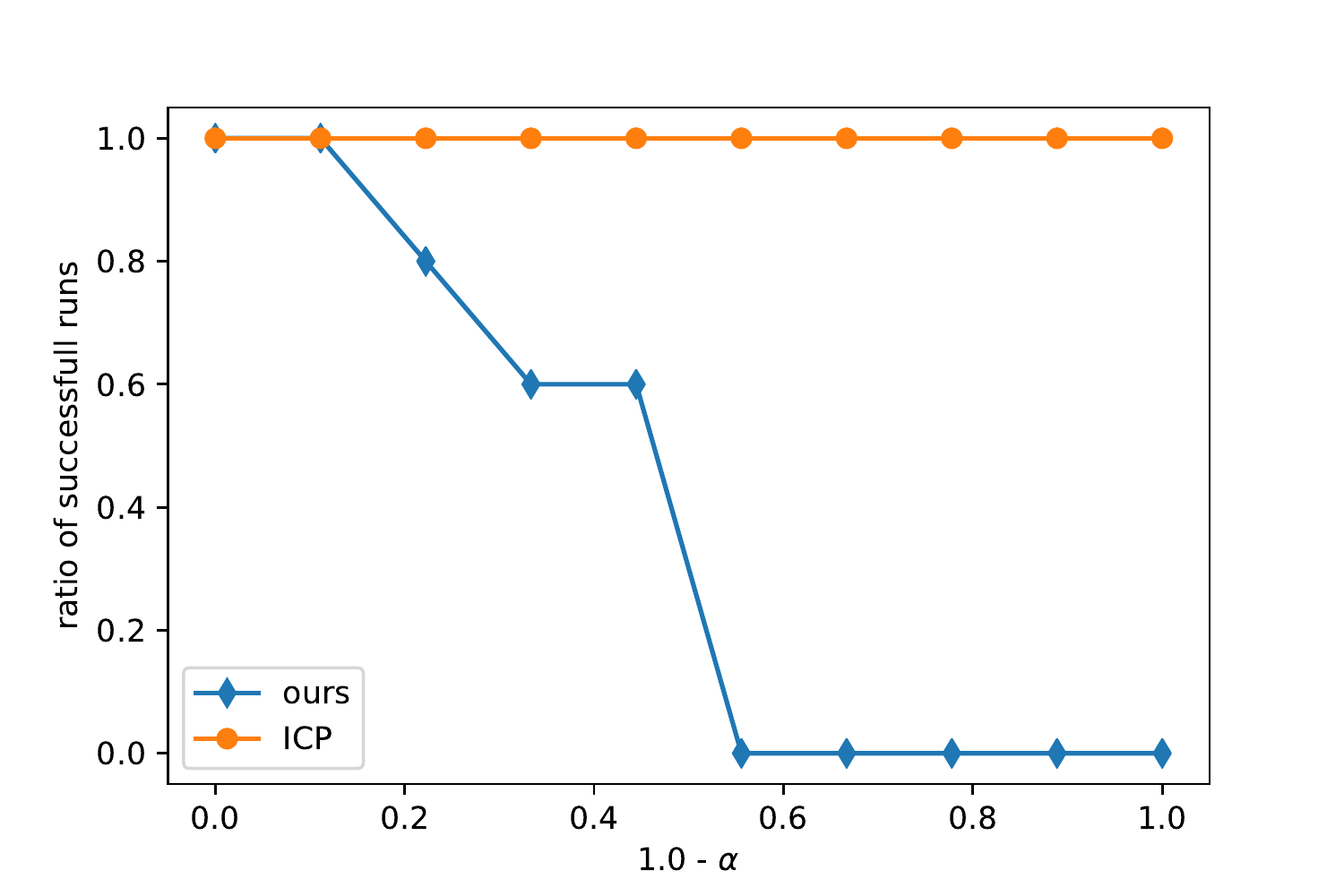}
    \caption{Success ratio overlap percentage}
    \label{fig:5}
\end{figure}

\begin{figure}[H]
    \centering
    \includegraphics[width=\linewidth]{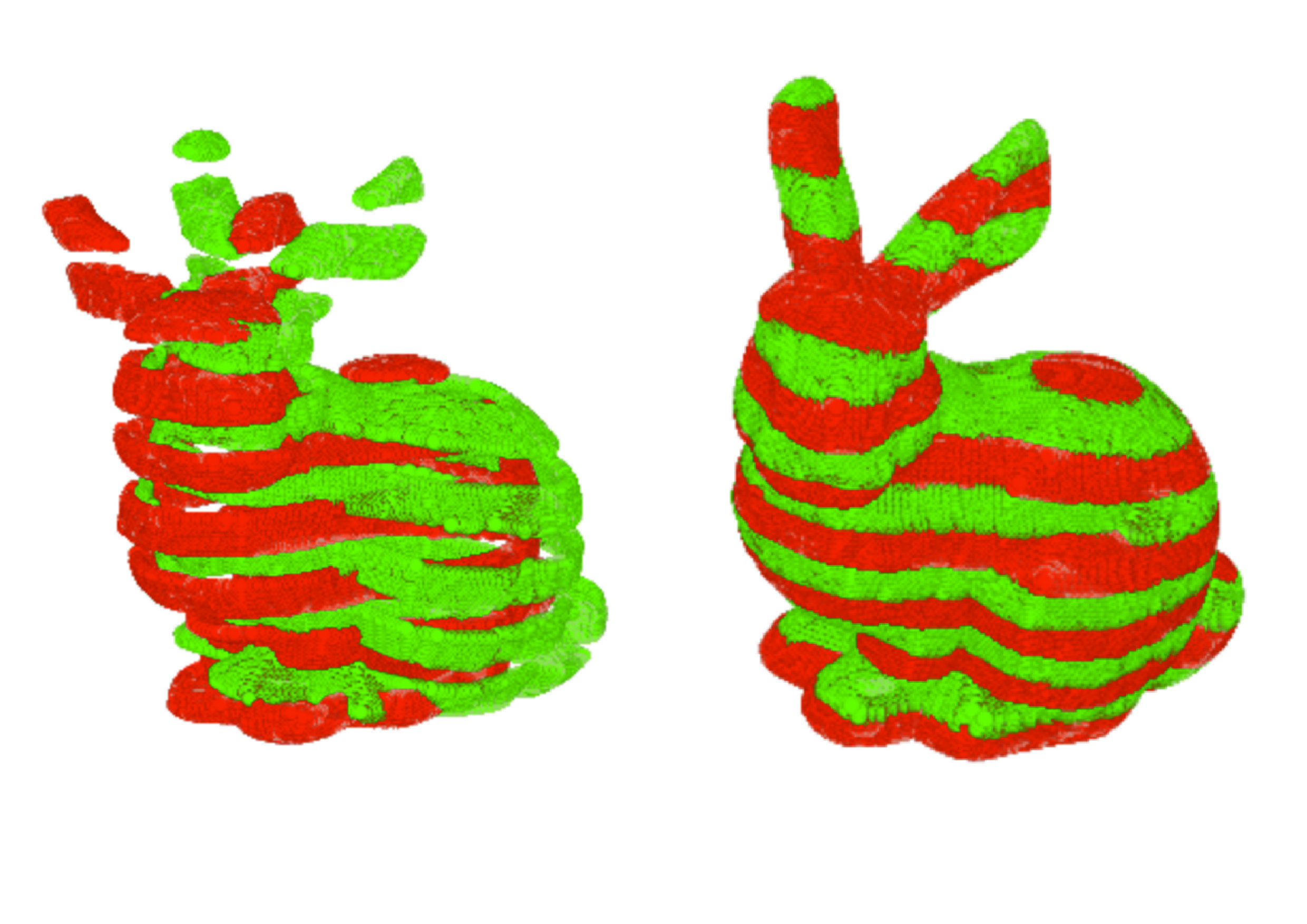}
    \caption{An example partial overlap augmentation.}
    \label{fig:6}
\end{figure}

We slice the mesh into a set of partially overlapping surfaces, with $\alpha$ - the percentage of overlap. We measure the success rate as a function of $\alpha$. Figure \ref{fig:6} demonstrates an example of a model sliced in the described way.

\subsubsection{\bf{Outliers}}

\begin{figure}[H]
    \centering
    \includegraphics[width=\linewidth]{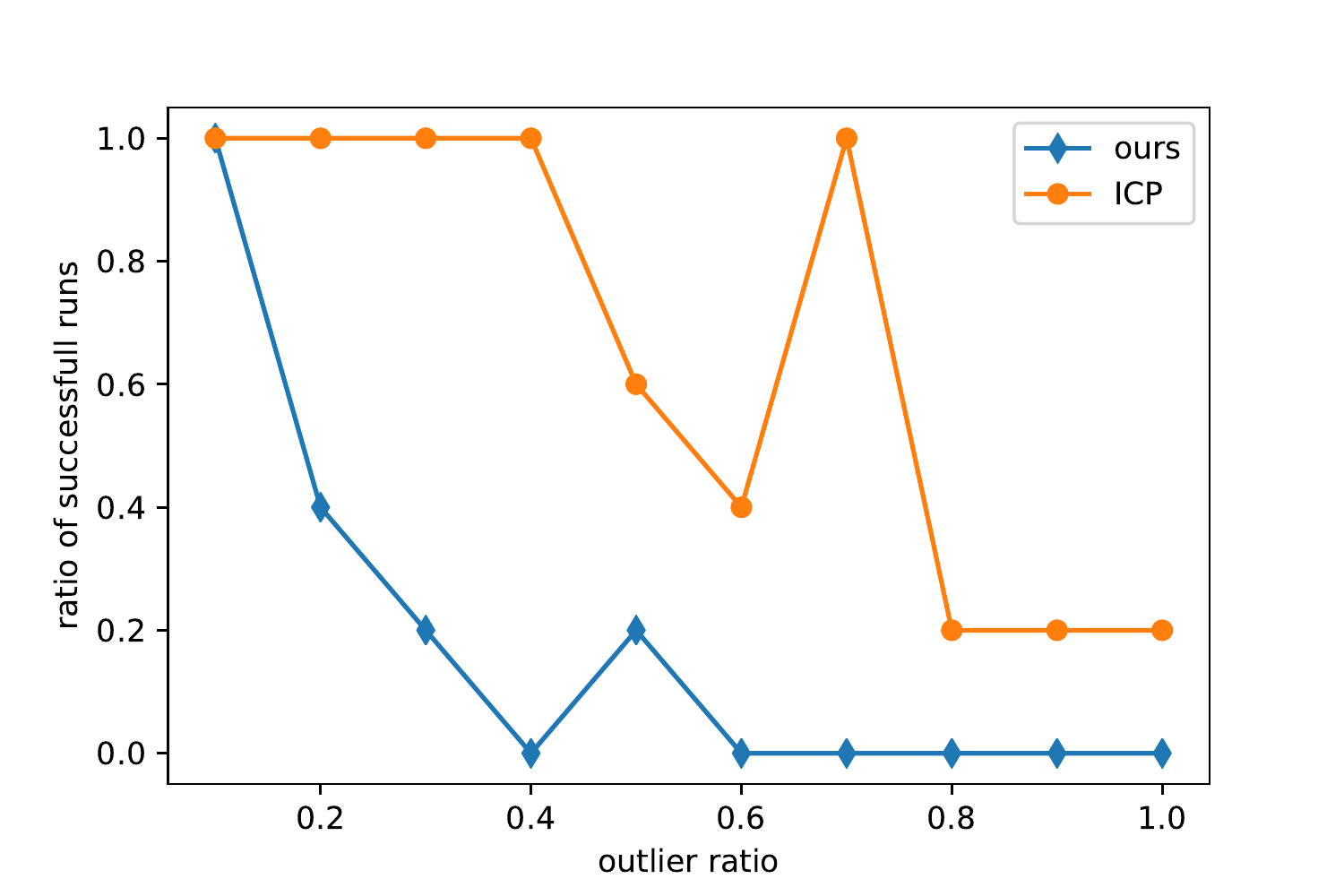}
    \caption{Success ratio to outlier percentage}
    \label{fig:7}
\end{figure}

\subsubsection{\bf{Outliers}}

We corrupt the source and target point sets with random points, uniformly distributed in the cube, containing the point set. Then we measure the performance with respect to the percentage of such outliers.

\subsubsection*{\bf{Discussion}}

The experiments show that our algorithm outperforms ICP in the ability to converge from severe initial rotations. Also, the algorithm turns out to be more robust to noise but fails to work when a large number of outliers is present.

\subsection{\bf{Global matching}}

To test the algorithm in the global registration setting, we have performed a real-life experiment, which was of practical importance to us. We have captured a dense depth image of a car using a high-resolution structured-light-stereo camera and tried to match it against a CAD model. What makes this problem especially challenging is that the nature of the data in source and target point clouds is very different. The number of points in the point cloud from the camera exceeds 400000, while the triangulated CAD model only contained around 5000 vertices. Because of such an aggressive optimization, the geometry of the CAD model differed significantly from that of the real point cloud, making the registration problem very ill-posed. The ground-truth was obtained by manual alignment.

\begin{figure}[H]
    \centering
    \includegraphics[width=\linewidth]{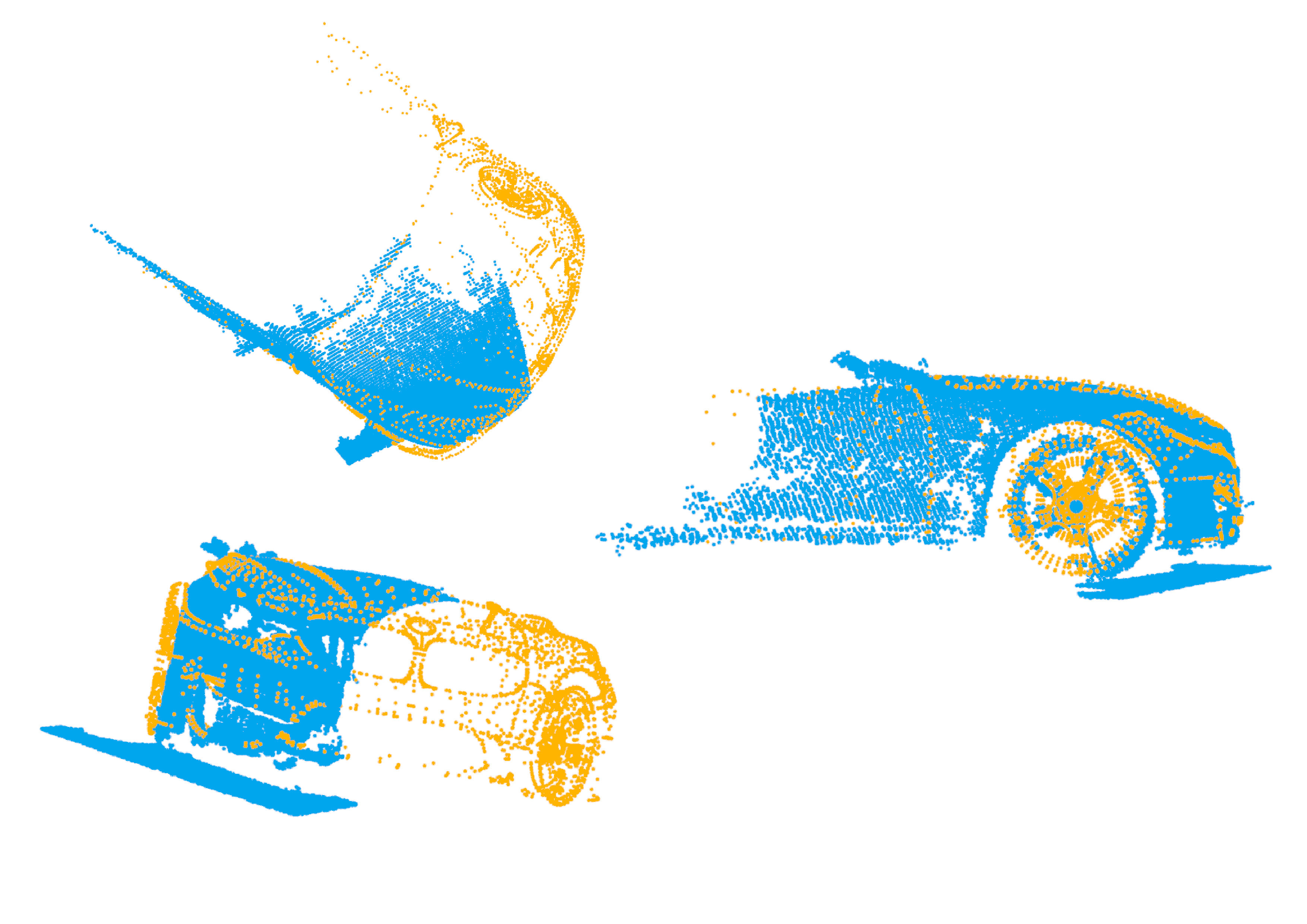}
    \caption{Result of our algorithm}
    \label{fig:8}
\end{figure}

This problem turned out to be quite challenging both for FPFH matching and for our method. Fig.\ref{fig:8} shows the result obtained by our algorithm. The rotation differs from ground-truth by 1.8 degrees. It takes our algorithm about 2000 epochs over the down-sampled point cloud with 7000 points to achieve such quality.

\begin{figure}[H]
    \centering
    \includegraphics[width=\linewidth]{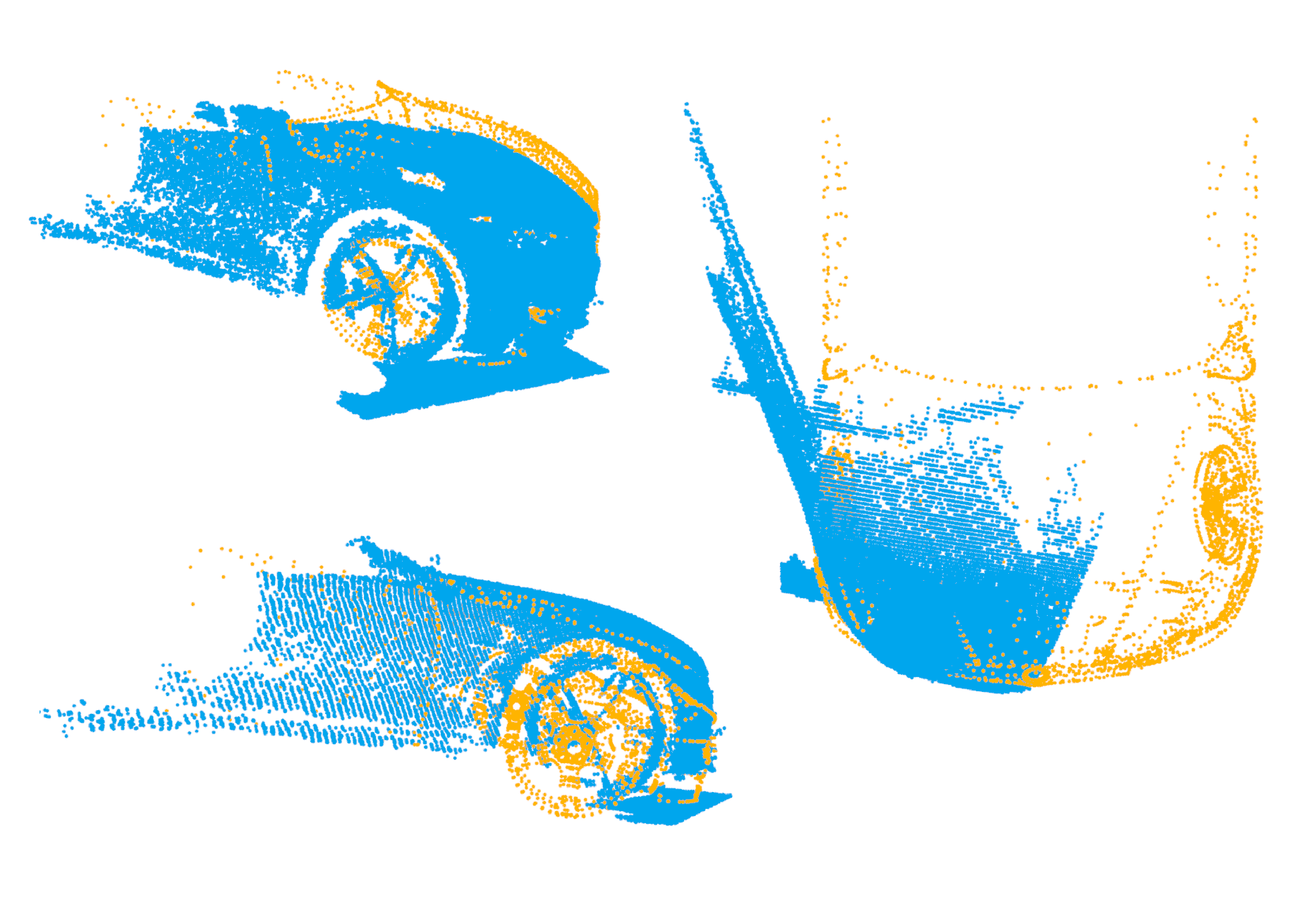}
    \caption{Results of FPFH matching}
    \label{fig:9}
\end{figure}

Computing FPFH descriptors requires to perform a down-sampling on the point clouds. It is done by voxelizing the space and taking one point from each voxel. The size of the voxel thus becomes the main parameter for FPFH construction. We tried running the FPFH matching with several voxel sizes ranging from 0.01 to 0.1 meters. Some of the resulting alignments are displayed in Fig. \ref{fig:9}

The smallest rotation error we've managed to obtain is 8 degrees (top left image), this corresponded to a voxel size of 0.08 meters.

\subsection{Learning translations}

\begin{figure}[H]
    \centering
    \includegraphics[width=\linewidth]{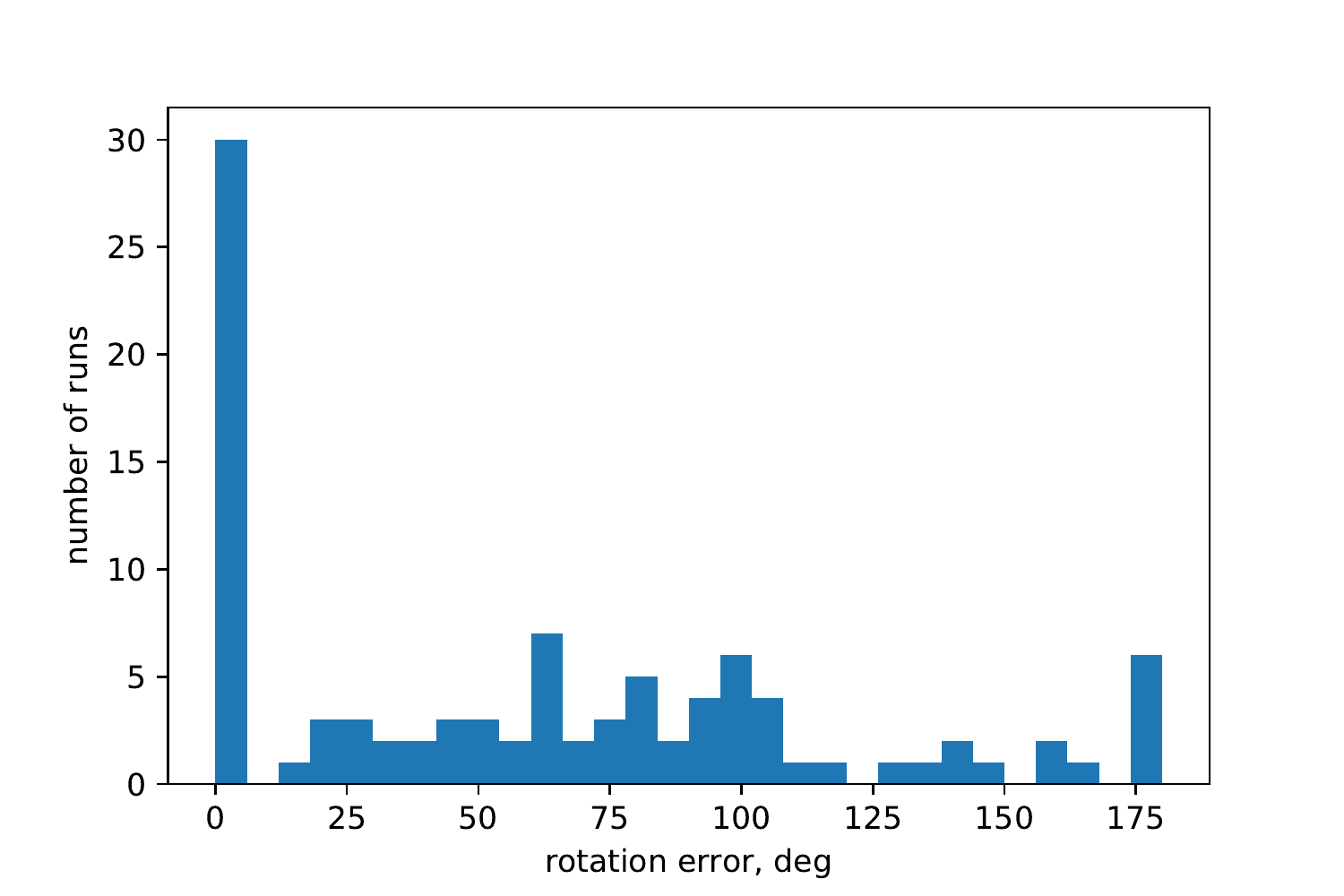}
    \caption{Rotation error distribution - learning rotation jointly with translation}
    \label{fig:10}
\end{figure}

The current version of our algorithm manages to stably find the rotation between the point sets, given that the translation is known. In this subsection, we discuss the behaviour of the algorithm when the translation is unknown. On Fig. \ref{fig:10} we show the distribution of resulting errors for a setup, where rotation and translation were jointly optimized. True rotation vectors were sampled from the standard normal distribution.

\begin{figure}[H]
    \centering
    \includegraphics[width=\linewidth]{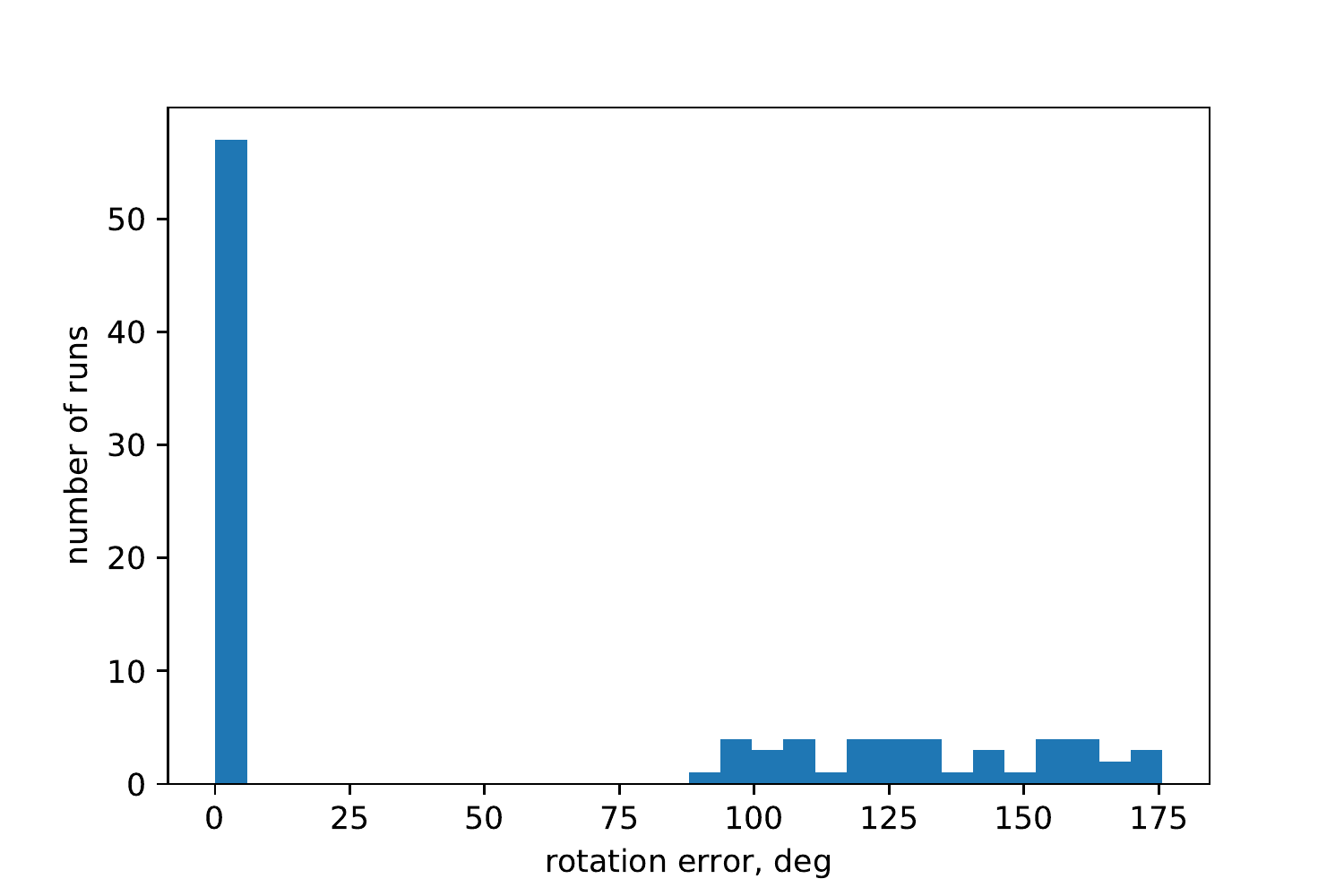}
    \caption{Rotation error distribution - learning only rotation with shift augmentations}
    \label{fig:11}
\end{figure}

In Fig. \ref{fig:11} we show the distribution of errors for the case, then the rotation is optimizing independently, prior to optimizing over translation. Here we employ the trick described in section 2.5 and observe that it does have a positive influence on the quality of the solution.

\section{Related work}
Iterative-closest point, originally introduced in \cite{121791} is without a doubt the most well-known algorithm for point set registration, which paved way for a large number of different modifications. ICP iterates over assigning correspondences between points of the target and source point clouds and performing the least squares optimization of the distances between corresponding points over the transformation parameters. These iterations continue until convergence.

Algorithms of this class suffer from a serious drawback - the need to find correspondences between the points of target and source point-clouds. The fact that the error function optimized by the ICP depends on hard correspondence assignment makes it a very ill-posed non-convex optimization problem, which possesses numerous local minima. Therefore, ICP-based algorithms are only to be used in cases where a reasonable approximation of the pose is already known. There also exists a modification of the ICP, which strives to find the global optimum using the Branch-and-Bound technique. \cite{DBLP:journals/corr/YangLCJ16}.

Another important development in this field was introduced in \cite{5432191}. It is a probabilistic method called Coherent Point Drift. Instead of relying on hard correspondence assignment like ICP, it actually matches each point of the source point cloud with all points from the target, by interpreting the target points as Gaussian Mixture Model centroids and performing the Expectation Maximization algorithm to find the transformation parameters, which maximize the likelihood of the source point-cloud. Coherent point drift is extremely computationally expensive, because of the posterior matrix computation. This can, however, be remedied by using Fast Gaussian Transform \cite{doi:10.1137/0912004}.

One more important research direction are so-called global matching algorithms, which aim to find a good starting point for ICP. A prominent class of methods in this area of research are Point Feature Histograms \cite{Rusu:2009:FPF:1703435.1703733}. Briefly, they work in the following way: for each point in both of the sets, a feature vector is computed, which captures the geometric information about a certain neighborhood of that point. The size of the neighborhood is a hyper-parameter of the algorithm. Then, global correspondences between points are built based on similarity of their feature vectors. Random Sample Consensus (RANSAC) algorithm is run to find the best correspondence in terms of the number of outliers.

\section{Conclusion}
In this work, we have demonstrated a novel approach to point set registration, which does not require matching geometrical features between point clouds. It highlights a new application for adversarial learning setups, showing that in some cases, the training procedure can be robust enough to be used for one-shot optimization. Although the large running time does not allow the algorithm to be used in real-time applications, the rapid development, which is now taking place in the field of adversarial networks is likely to change that soon.


{\small
\bibliographystyle{ieee}
\bibliography{egbib}
}

\end{document}